\documentclass[a4paper]{article}
\usepackage{eacl2021}
\usepackage{times}
\usepackage{latexsym}

\usepackage{microtype}
\usepackage{graphicx}
\usepackage{subcaption}
\usepackage{multirow}
\usepackage{adjustbox}
\usepackage{amsmath}
\usepackage{subcaption}
\usepackage{enumitem}
\usepackage{graphics}
\usepackage{amsfonts}
\usepackage{mathrsfs}
\usepackage{amssymb}
\usepackage{amsthm}
\usepackage{fixmath}
\usepackage{tikz} 
\usepackage{float}
\usepackage{tabularx}
\usepackage{caption}
\usepackage{comment}
\usepackage{threeparttable}
\usepackage{siunitx}
\restylefloat{table}
\newcommand{\xhdr}[1]{{\noindent\bfseries #1}.}
\definecolor{ultramarine}{RGB}{255,105,180}
\definecolor{steelblue}{RGB}{51, 117, 204}
\definecolor{skyblue}{RGB}{102, 157, 229}
\definecolor{lightblue}{RGB}{153,197,255}

\aclfinalcopy % Uncomment this line for the final submission
\title{Selective Attention Based Graph Convolutional Networks for Aspect-Level Sentiment Classification}

\author{Xiaochen Hou\textsuperscript{$*$}, Jing Huang, Guangtao Wang\\ \textbf {Xiaodong He, Bowen Zhou} \\
\\
\large{JD AI Research, Mountain View, California}\\
  {\tt \textsuperscript{$*$}xiaochen.hou1@jd.com} \\
}
\date{}

\begin{document}
\maketitle
\begin{abstract}
Aspect-level sentiment classification aims to identify the sentiment polarity towards a specific aspect term in a sentence. 
% Recent approaches employ Graph Convolutional Networks (GCN) over dependency trees to shorten the distance between the aspect term and opinion words and benefit from syntactic relationships between them.
Recent approaches employ Graph Convolutional Networks (GCN) over dependency trees to obtain syntax-aware representations of aspect terms and learn interactions between aspect terms and context words. GCNs often achieves the best performance with two layers and deeper GCNs do not bring any additional gain. However, in some cases, 
%important context 
the corresponding opinion words for an aspect term cannot be reached within two hops on dependency trees. 
% However, the GCN model over dependency trees alone is vulnerable to parsing errors.
% In order to alleviate problems caused by parsing errors, 
% a straightforward solution would just combine a self-attention sequence model with the GCN model. 
%since a sequence model allows direct interactions between the aspect term and all other words in the sequence. 
% Instead of naively taking an ensemble of these two models, 
% However, due to the dependency parsing errors and complex syntactic structure of a sentence, an aspect term could still be far away from the opinion words on the dependency tree. Deeper GCNs cannot handle this situation effectively according to previous works. 
Therefore, we design a novel selective attention based GCN model (\textit{SA-GCN}) to handle the situation where aspect terms are far away from opinion words. 
Because opinion words are direct explanation for the aspect-term polarity classification, we use the opinion extraction as an auxiliary task to help the sentiment classification task.
% for joint aspect-term sentiment classification and opinion extraction.
Specifically, on top of the GCN model operating on the dependency tree, we use the self-attention to directly select $k$ words with highest attention scores for each word in the sentence. Then we apply another GCN model on the generated top $k$ attention graph to integrate the information from selected context words. 
% Jing: depending on the experimental results, edit this sentence later
% The joint training benefits both aspect-term sentiment classification as well as opinion extraction. 
We conduct experiments on 
several commonly used benchmark datasets.
The experiments show that our proposed \textit{SA-GCN} achieves new state-of-the-art results. 
%on the SemEval datasets.
\end{abstract}

\section{Introduction}

Aspect-level sentiment classification is a fine-grained sentiment analysis task, which aims to identify the sentiment polarity (e.g., positive, negative or neutral) of a specific aspect term (also called target) appearing in a review. For example, ``\textit{Despite a slightly limited menu, everything prepared is done to perfection, ultra fresh and a work of food art.}'', the sentiment polarity of aspect terms ``menu'' and ``food'' are negative and positive, respectively. And the opinion words ``limited'' and ``done to perfection'' provide evidences for sentiment polarity predictions. This task has many applications, such as restaurant recommendation and purchase recommendation on e-commerce websites.

To solve this problem, recent studies have shown that the interactions between an aspect term and its context (which include opinion words) are crucial to identify the sentiment polarity towards the given term. Most approaches consider the semantic information from the context words and utilize the attention mechanism to learn such interactions. However, it has been shown that syntactic information obtained from dependency parsing is very effective in capturing long-range syntactic relations that are obscure from the surface form~\cite{zhang2018graph}. A recent popular approach to learn syntax-aware representations is employing graph convolutional networks (GCN)~\cite{kipf2016semi} model over dependency trees~\cite{huang2019syntax,zhang2019aspect,sun2019aspect,wang2020relational,tang-etal-2020-dependency}, which allows the message passing between the aspect term and its context words in a syntactical manner. 

\begin{figure}[h]
\centering
\includegraphics[width=0.95\linewidth]{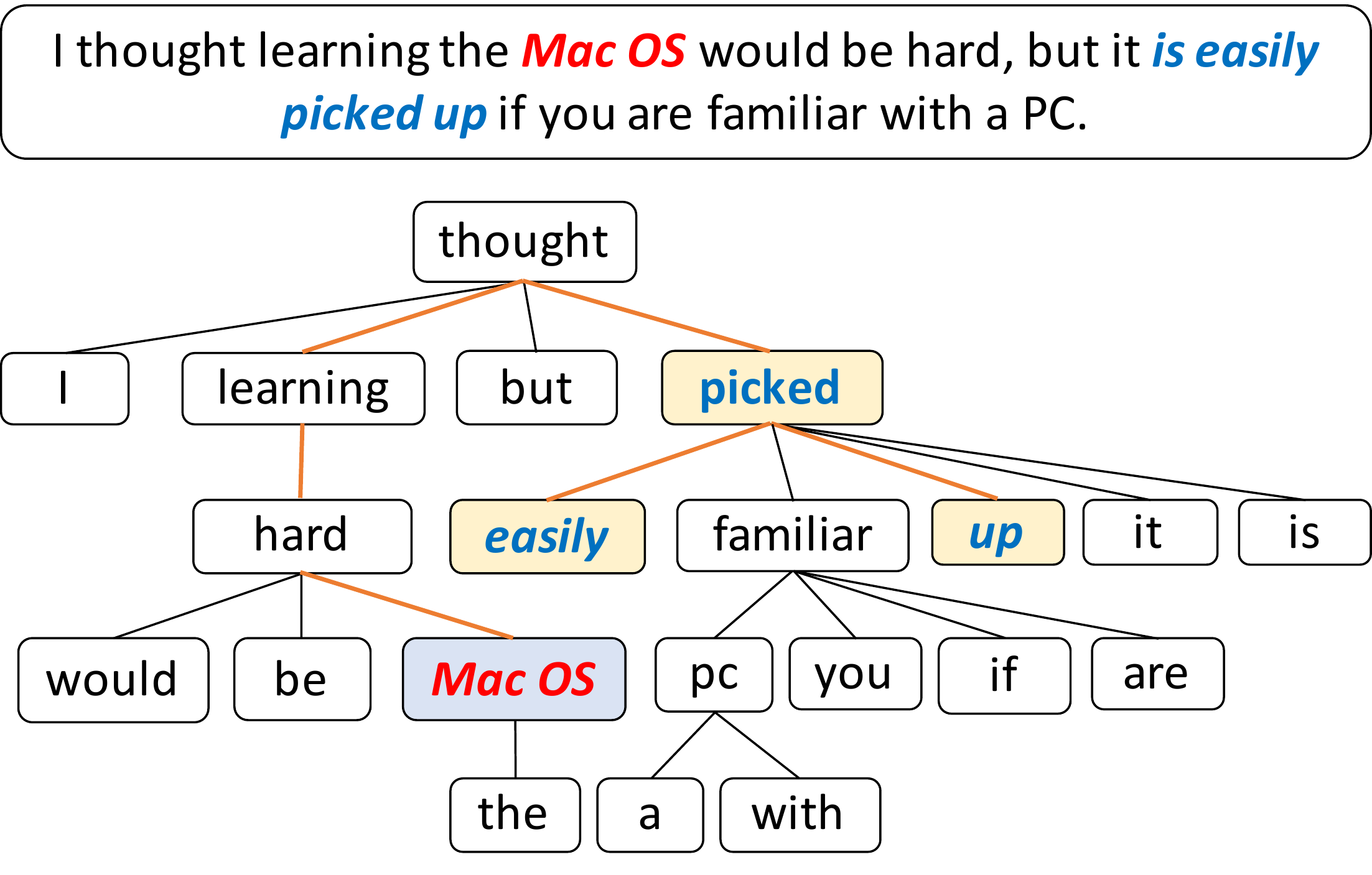}
\caption{Example of dependency tree with multi-hop between aspect term and determined context words.}
\label{fig:dtl}
\end{figure}

Previous works show that GCN models with two layers achieve the best performance~\cite{zhang2018graph,xu2018representation}. Deeper GCNs do not bring additional gain due to the over-smoothing problem~\cite{li2018deeper}, which makes different nodes have similar representations and lose the distinction among nodes. However, in some cases, the most important context words, i.e. opinion words, are more than two-hops away from the aspect term words on the dependency tree.
% However, the prediction results are highly dependent on the accuracy of the dependency parser.
% Parsing errors could make the aspect term and opinion words far away from each other and introduce noisy context for layers in GCN. 
As indicated by Figure \ref{fig:dtl}, there are four hops between the target ``Mac OS'' and the opinion words ``easily picked up'' on the dependency tree.
% compared with the correct parsing tree \ref{fig:correct}, 
% the dependency tree \ref{fig:wrong} 
% mistakenly separate the two parts, ``I had the soup'' and ``it was not tasty at all'', 
% Although using deeper GCNs could still pass information among distant nodes, most of previous work already indicated that GCN models with two layers achieved the best performance~\cite{kipf2016semi,xu2018representation,zhang2019aspect,sun2019aspect} and more layers of GCNs did not bring additional gain~\cite{li2018deeper,wang2019improving} due to noise introduced by distant nodes. 

% However, the long-distance problem, that refers to the situation where aspect terms are more than two-hops away from the opinion words on dependency trees, 
% is quite common and usually caused by two reasons. First, the possible dependency parsing errors could separate the aspect term far away from the opinion words. As indicated by Figure \ref{fig:wd}Second, the complicated syntactic structure of a sentence could make aspect terms and opinion words far apart. Figure \ref{fig:dtl} demonstrates a correct dependency tree, but because of the complexity of its syntactic structure, the dependency tree does not help with shortening the distance between the aspect term ``Mac OS'' and the opinion words ``easily picked up''.

In order to solve the above problem, 
we propose a novel selective attention based GCN (\textit{SA-GCN}) model that
% a straightforward solution is 
combines the GCN model over dependency trees with a self-attention based sequence model. The self-attention sequence model enables the direct interaction between an aspect term and its opinion words so that it can take care of the situation where the term is far away from the opinion words on the dependency tree. 
% Nevertheless, a naive ensemble of these two models obtains very small performance gains over each individual model (see Table~\ref{table:results}).

% Therefore, in this paper, we propose a novel selective attention based GCN (\textit{SA-GCN}) model, which combines the benefits of GCN model over dependency trees and a self-attention sequence model. 
%which is robust to handle the long-distance problem by allowing the aspect term to get direct access to the opinion words. 
Specifically, the base model is the GCN model over dependency trees, which applies the pre-trained BERT as an encoder to obtain representations of the aspect term and its context words as the initial node features on the dependency tree. 
This model considers the connections between the target and its syntactic neighbors on the dependency tree.
%This GCN baseline module fuses the syntactic knowledge presented in the dependency tree and semantic information from the BERT encoder. 

Next, the GCN outputs are fed into a top-$k$ multi-head attention selection module. For each head, top-$k$ important context words are selected according to the attention score matrix. This step of selection effectively removes noisy and unrelated words from the context for the aspect term.
Then on top of the selected attention score matrix which represents a new graph, we apply a GCN layer again to integrate information from the top-$k$ important context words. Therefore, the final aspect term representation integrates semantic representation from BERT, syntactic information from the dependency tree, and the top-$k$ attended context words from the sentence sequence. This representation is then fed into the final classification layer for sentiment prediction.

%\jing{please add a paragraph for motivation of joint training of sentiment classification and opinion extraction}
We further enhance the training of sentiment classification with an auxiliary task of opinion extraction. The intuition is that locating opinion words for the aspect term could benefit the prediction of sentiment polarity. As shown in Figure \ref{fig:dtl}, if the opinion words ``easily picked up'' are detected correctly, it definitely could help the model to classify the sentiment as positive. 
In fact, our top-$k$ selection module is designed to find such opinion words. Therefore, we further introduce the opinion words extraction task to provide supervision information for the top-$k$ selection procedure. In details, we directly feed the \textit{SA-GCN} output to a CRF decoder layer, and jointly train the sentiment classification and opinion extraction tasks.

The main contributions of this work are summarized as the following:
\begin{itemize}[leftmargin=6pt]
\setlength{\itemsep}{0pt}%
    \setlength{\parskip}{2pt}
    \item %The drawback of the GCN model alone over dependency trees is that it is not able to handle the long distance between the aspect term and opinion words. To solve this problem, 
    We propose a selective attention based GCN (\textit{SA-GCN}) module, which takes the benefit of GCN over the dependency trees and enables the aspect term directly obtaining information from the opinion words according to most relevant context words. This helps the model handle cases when the aspect term and opinion words are located far away from each other on the dependency tree.
    \item We propose to jointly train the sentiment classification and opinion extraction tasks. The joint training further improves the performance of the classification task and provides explanation for sentiment prediction.
    \item We conduct experiments on four benchmark datasets including Laptop and Restaurant reviews from SemEval 2014 Task 4, Restaurant reviews from SemEval 2015 Task 12 and SemEval2016 Task 5, and our \textit{SA-GCN} achieves new state-of-the-art results.
\end{itemize}

\section{Related Work}
Capturing the interaction between the aspect term and opinion words is essential for predicting the sentiment polarity towards the aspect term. In recent work, various attention mechanisms, such as co-attention, self-attention and hierarchical attention, were utilized to learn this interaction~\cite{tang2015effective,tang2016aspect,liu2017attention,li2018transformation,wang2018target,fan2018multi,chen2017recurrent,zheng2018left,wang2018learning,li2018hierarchical,li2018transformation}. Specifically, they first encoded the context and the aspect term by recurrent neural networks (RNNs), and then stacked several attention layers to learn the aspect term representations from important context words.

After the success of the pre-trained BERT model~\cite{devlin2018bert},~\citet{song2019attentional} utilized the pre-trained BERT as the encoder.
In the study by~\cite{xu2019bert}, the task was considered as a review reading comprehension (RRC) problem. RRC datasets were post trained on BERT and then fine-tuned to the aspect-level sentiment classification.~\citet{rietzler2019adapt} utilized millions of extra data based on BERT to help sentiment analysis.

The above approaches mainly considered the semantic information. Recent approaches attempted to incorporate the syntactic knowledge to learn the syntax-aware representation of the aspect term.~\citet{dong2014adaptive} proposed AdaRNN, which adaptively propagated the sentiments of words to target along the dependency tree in a bottom-up manner.~\citet{nguyen2015phrasernn} extended RNN to obtain the representation of the target aspect by aggregating the syntactic information from the dependency and constituent tree of the sentence.~\citet{he2018effective} proposed to use the distance between the context word and the aspect term along the dependency tree as the attention weight. Some researchers ~\cite{huang2019syntax,zhang2019aspect,sun2019aspect} employed GNNs over dependency trees to aggregate information from syntactic neighbors. Most recent work in ~\citet{wang2020relational} proposed to reconstruct the dependency tree to an aspect-oriented tree. The reshaped tree only kept the dependency structure around the aspect term and got rid of all other dependency connections, which made the learned node representations not fully syntax-aware. ~\citet{tang-etal-2020-dependency} designed a mutual biaffine module between Transformer encoder and the GCN encoder to enhance the representation learning.

The downside of applying GCN over dependency trees is that it cannot elegantly handle the long distance between aspect terms and opinion words.
% \footnote{In \cite{huang2019syntax} 5-layer GAT was used, because GAN outputs were not directly used for classification rather being passed through a LSTM model.}
Our proposed \textit{SA-GCN} model effectively integrates the benefit of a GCN model over dependency trees and a self-attention sequence model to directly aggregate information from opinion words.
The top-$k$ self-attention sequence model selects the most important context words, which effectively sparsifies the fully-connected graph from self-attention. Then we apply another GCN layer on top of this new sparsified graph, 
% aggregate information from them by another GCN layer, 
such that each of those important context words is directly reachable by the aspect term and the interaction between them could be learned. 
%In addition, we incorporate the opinion extraction as an auxiliary task to further guide the learning of top-$k$ selection and improve the sentiment classification performance.

% The GCN model on the dependency tree and the \textit{SA-GCN} model on the selected attention graph are trained together to obtain synergistic results.

\section{Proposed Model}
% In this section, we first present the overview of our model. Then, we introduce the main modules in our model in details.
%In our model, each training instance is composed of a sentence-term pair, referring to a sentence and an aspect term appearing in the sentence. The goal of this model is to predict the sentiment polarity of the aspect term. Figure \ref{fig:model} illustrates the overall architecture of the proposed model.

% In this section, we introduce the proposed model in detail. We first introduce the general view of our model. Then, we introduce the main modules in our models in details.

%In our model, each training instance is composed of a sentence-term pair, referring to a sentence and an aspect term appearing in the sentence. The goal of this model is to predict the sentiment polarity of the aspect term. Figure \ref{fig:model} illustrates the overall architecture of the proposed model.

\subsection{Overview of the Model}
\begin{figure*}[!h]
\centering
\includegraphics[width=0.9\linewidth]{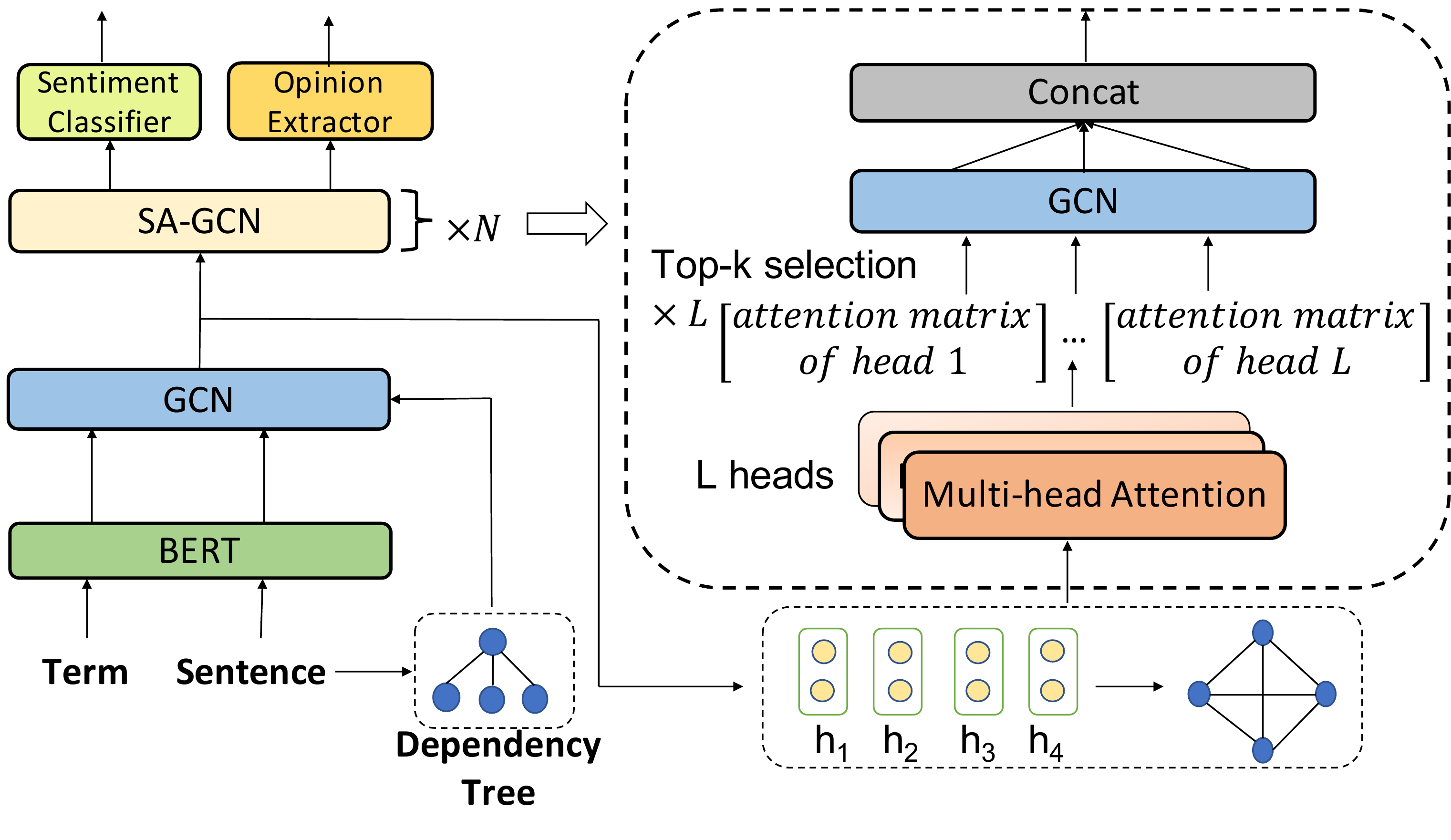}
\caption{The \textit{SA-GCN} model architecture: the left part is the overview of the framework, the right part shows details of a \textit{SA-GCN} block.}
\vspace{-10pt}
\label{fig:model}
\end{figure*}

The goal of our proposed \textit{SA-GCN} model is to predict the sentiment polarity of an aspect term in a given sentence. To improve the sentiment classification performance and provide explanations about the polarity prediction, we also introduce the opinion extraction task for joint training. The opinion extraction task aims to predict a tag sequence $\mathbold{y}_{o} = [y_1, y_2, \cdots, y_n]$ ($y_{i}\in \{B, I, O\}$) denotes the beginning
of, inside of, and outside of opinion words.
Figure \ref{fig:model} illustrates the overall architecture of the \textit{SA-GCN} model.
For each instance composing of a sentence-term pair, all the words in the sentence except for the aspect term are defined as context words. 
% As illustrated in Figure \ref{fig:model}, we perform the aspect sentiment classification by the following steps: (1) encode both the aspect terms and context words by BERT, and use these representations as the initial features of the nodes (i.e., either context words or the aspect term) in the dependency tree; (2) perform GCN over the dependency tree of the sentence; (3) employ a novel selective attention based GCN (See the right part in Figure \ref{fig:model}) to learn the representation of the aspect term; (4) make the sentiment prediction and opinion extraction based on the aspect term's representation induced from former steps.

\subsection{Encoder for Aspect Term and Context}
\xhdr{BERT Encoder}
We use the pre-trained BERT base model as the encoder to obtain embeddings of sentence words. Suppose a sentence consists of $n$ words $\{w_1,w_2,...,w_{\tau}, w_{\tau+1}...,w_{\tau+m}, ...,w_n\}$ where $\{w_{\tau}, w_{\tau+1}...,w_{\tau+m-1}\}$ stand for the aspect term containing $m$ words. First, we construct the input as ``[CLS] + sentence + [SEP] + term + [SEP]'' and feed it into BERT. This input format enables explicit interactions between the whole sentence and the term such that the obtained word representations are term-attended.
Then, we use average pooling to summarize the information carried by sub-words from BERT and obtain final embeddings of words $\mathbold{X} \in \mathbb{R}^{{n} \times d_{B}}$, $d_B$ refers to the dimensionality of BERT output. 
% Similarly, term representation $\mathbold{X}_{t} \in \mathbb{R}^{m \times d_{B}} $ is obtained, where $d_{B}$ is the dimension of the BERT output.

% \xhdr{Self-attention layer:}
% after obtaining the embedding of the aspect term, we apply self-attention to summarize the information carried by each sub-token of the aspect term and get a single feature representation as the term feature~\cite{zhong2019coarse}. We utilize a two-layer Multi-Layer Perceptron (MLP) to compute the scores of sub-tokens and get weighted sum over all sub-tokens. This is formulated as follows:
% \begin{flalign}
% \mathbold{a}&=softmax({\sigma({\mathbold{W}_{2}\sigma({\mathbold{W}_{1}\mathbold{X}_{t}^T})})})\\
% \mathbold{h}_{a}&=\mathbold{a}\mathbold{X}_{t}
% \end{flalign}
% where $\mathbold{a}\in \mathbb{R}^{1 \times m}$, $\mathbold{h}_{a} \in \mathbb{R}^{1 \times d_{B}}$, $\mathbold{X}_t^T$ is the transposition of $\mathbold{X}_t$, and $\sigma$ denotes $\tanh$ activation function. The bias vectors are not shown here for simplicity.
% % \begin{figure}
% % \centering
% % \includegraphics[width=\linewidth]{dt.pdf}  
% % \caption{Example of Dependency Tree}
% % \label{fig:dt}
% % \end{figure}
\subsection{GCN over Dependency Trees}
With 
% the aspect term representation $\mathbold{h}_{a}$ and 
words representations $\mathbold{X}$ as node features and dependency tree as the graph, we employ a GCN to capture syntactic relations between the term node and its neighboring nodes. 
% An example of the dependency tree is presented in Figure \ref{fig:correct}. With the correct dependency tree, the aspect term ``soup'' is connected with the sentiment context ``tasty'' via two hops. 

%GCNs are designed to deal with data containing graph structure. A graph is constructed by nodes and edges.
GCNs have been shown to be effective models for many NLP applications, such as relation extraction~\cite{guo2019attention,zhang2018graph}, reading comprehension~\cite{kundu2018exploiting,tu2019hdegraph}, and aspect-level sentiment analysis~\cite{huang2019syntax,zhang2019aspect,sun2019aspect}.
In each GCN layer, a node aggregates the information from its one-hop neighbors and update its representation. 
%If two GCN layers are used, the above process is repeated twice, so that each node gets information from two-hop away neighbors. 
In our case, the graph is represented by the dependency tree, where each word is treated as a single node and its representation is denoted as the node feature. 
The message passing on the graph can be formulated as follows:
\begin{flalign}
\mathbold{H}^{(l)}&=\sigma(\mathbold{A}\mathbold{H}^{(l-1)}\mathbold{W})
\end{flalign}
where $\mathbold{H}^{(l)} \in \mathbb{R}^{n \times d_h}$ is the output $l$-th GCN layer, $\mathbold{H}^{(0)} \in \mathbb{R}^{n \times d_B}$ is the input of the first GCN layer, and $\mathbold{H}^{(0)}=\mathbold{X} \in \mathbb{R}^{n \times d_B}$.
% and the aspect term  $\mathbold{h}_{a} \in \mathbb{R}^{1 \times d_B}$. 
$\mathbold{A} \in \mathbb{R}^{n \times n}$ denotes the adjacency matrix obtained from the dependency tree, note that we add a self-loop on each node. $\mathbold{W} \in \mathbb{R}^{d_B \times d_h}$ represents the learnable weights and $\sigma$ refers to $ReLU$ activation function.

The node features are passed through the GCN layer, the representation of each node is now further enriched by syntax information from the dependency tree.
 
\subsection{SA-GCN: Selective Attention based GCN}
% Although performing GCN over the dependency trees could help to shorten the distance between the aspect term and opinion words, there are also some issues caused by parsing errors. For example, 
% aspect term and opinion words are made further apart due to dependency parsing errors as indicated in Figure \ref{fig:wd}, and the noisy context is used in GCN layers.
Although performing GCNs over dependency trees brings syntax information to the representation of each word, it also limits interactions between aspect terms and long-distance opinion words that are essential for determining the sentiment polarity.
% Performing GCN over the dependency trees alone is vulnerable to parsing errors, which could make aspect term and opinion words further apart and introduce noisy context.
In order to alleviate the problem, we apply a Selective Attention based GCN (\textit{SA-GCN}) block to identify the most important context words and integrate their information into the representation of the aspect term. Multiple \textit{SA-GCN} blocks can be stacked to form a deep model. 
Each \textit{SA-GCN} block is composed of three parts: a multi-head self-attention layer, top-$k$ selection and a GCN layer. 
% We will introduce them in detail in the following sections.

\xhdr{Self-Attention}
We apply the multi-head self-attention first to get the attention score matrices $\mathbold{A}_{score}^i\in \mathbb{R}^{n \times n}$($1\leq i\leq L$), $L$ is the number of heads. It can be formulated as:
\begin{flalign}
\mathbold{A}_{score}^i&=\frac{(\mathbold{{H}}_{k,i}\mathbold{{W}_{k}})(\mathbold{{H}}_{q,i}\mathbold{W_{q}})^T}{\sqrt{d_{head}}}\\
d_{head}&=\frac{d_h}{L}
\end{flalign}
where $\mathbold{H}_{*,i}=\mathbold{H}_{*}[:,:,i]$, $* \in \{k\text{: key}, q\text{: query}\}$, $\mathbold{H}_k \in \mathbb{R}^{n \times d_{head} \times L}$ and $\mathbold{H}_q \in \mathbb{R}^{n \times d_{head} \times L}$ are the node representations from the previous GCN layer, $\mathbold{W}_k \in \mathbb{R}^{d_{head} \times d_{head}} $ and $\mathbold{W}_q \in \mathbb{R}^{d_{head} \times d_{head}} $ are learnable weight matrices, $d_h$ is the dimension of the input node feature, and $d_{head}$ is the dimension of each head. 

% This step allows the aspect term to directly connected to the most important context words.
The obtained attention score matrices can be considered as $L$ fully-connected (complete) graphs, where each word is connected to all the other context words with different attention weights. This kind of attention score matrix has been used in attention-guided GCNs for relation extraction~\cite{guo2019attention}. Although the attention weight is helpful to differentiate 
different words, the fully connected graph still results in the aspect node fusing all the other words information directly, and the noise is often introduced during feature aggregation in GCNs, which further hurts the sentiment prediction. Therefore, we propose a top-$k$ attention selection mechanism to sparsify the fully connected graph, and obtain a new sparse graph for feature aggregation for GCN. This is different from attention-guided GCNs~\cite{guo2019attention} which performed feature aggregation over the fully-connected graph. Moreover, our experimental study (see Table \ref{table:ablation} in Section \ref{sec:experiments}) also confirms that the top-$k$ selection is quite important and definitely beneficial to the aspect-term classification task.

\xhdr{Top-$k$ Selection} For each attention score matrix $\mathbold{A}_{score}^i$, we find the top-$k$ important context words for each word, which effectively remove some edges in $\mathbold{A}_{score}^i$.
% add more explanations about why top-k is helpful for solving the long-distance problem.
The reason why we only choose the top-$k$ context words is that only a few words are sufficient to determine the sentiment polarity towards an aspect term. Therefore, we discard other words with low attention scores to get rid of irrelevant noisy words.
% The reason why we only choose the top-$k$ instead of keeping all the context words is that, some unimportant context words could introduce noise and cause confusion to the classification of the sentiment polarity. For example, the sentence is \textit{``To be completely fair, the only redeeming factor was the food, which was above average, but couldn't make up for all the other deficiencies of Teodora."}, the aspect term is \textit{``food''} and the sentiment label is positive. Without the top-$k$ selection, \textit{``food''} gets direct access to the context word \textit{``deficiencies''}, and it might result in classifying the polarity of \textit{``food''} to be negative. But if we only keep the crucial context words, such as \textit{``redeeming''} and \textit{`` above average''}, the potential risk could be eliminated. Thus we directly choose $k$ context words with the highest attention weights, and get rid of the probable noise brought by other context words. 

We design two strategies for top-$k$ selection, head-independent and head-dependent.
%global view and local view. 
% \todowork {Discuss the motivation and difference of two strategies}
Head-independent selection determines $k$ context words by aggregating the decisions made by all heads and reaches to an agreement among heads, while head-dependent policy enables each head to keep its own selected $k$ words.
% Head-independent selection sums up the attention score matrix of each head, such that the context words that are considered important by most heads get emphasized and context words that only show value in single head get ignored. 
% % enables information exchange among all attention heads. Specifically, it makes 
% Whereas, head-dependent selection finds the top $k$ context words for each attention score matrix individually, thus each head is able to focus on context words according to its own perspective. 

Head-independent selection is defined as following: we first sum the attention score matrix of each head element-wise, and then find top-$k$ context words using the mask generated by the function $topk$. For example, {\em topk}$([0.3,0.2,0.5])$ returns $[1,0,1]$ if $k$ is set to 2. Finally, we apply a softmax operation on the updated attention score matrix. The process could be formulated as follows:
\begin{flalign}
\mathbold{A}_{sum}&=\sum_{i=1}^{L}\mathbold{A}_{score}^i\\
\mathbold{A}_{m_{ind}}&=topk(\mathbold{A}_{sum})\\
\mathbold{A}_{h_{ind}}^i&=softmax(\mathbold{A}_{m_{ind}}\circ\mathbold{A}_{score}^i)
\end{flalign}
where $\mathbold{A}_{score}^i$ is the attention score matrix of $i$-th head, $\circ$ denotes the element-wise multiplication.

Head-dependent selection finds top-$k$ context words according to the attention score matrix of each head individually. We apply the softmax operation on each top-$k$ attention matrix. This step can be formulated as:
\begin{flalign}
\mathbold{A}^{i}_{m_{dep}}&=topk(\mathbold{A}^{i}_{score})\\
\mathbold{A}_{h_{dep}}^i&=softmax(\mathbold{A}^{i}_{m_{dep}} \circ \mathbold{A}_{score}^i)
\end{flalign}
Compared to head-independent selection with exactly $k$ words selected, head-dependent usually selects a larger number (than $k$) of important context words. Because each head might choose different $k$ words thus more than $k$ words are selected in total.
%and aggregates these words into the representation of the aspect-term.

From top-$k$ selection we obtain $L$ graphs based on the new attention scores and pass them to the next GCN layer. For simplicity, we will omit the $head$-$ind$ and $head$-$dep$ subscript in the later section. The obtained top-$k$ score matrix $\mathbold{A}$ could be treated as an adjacency matrix, where $\mathbold{A}(p,q)$ denotes as the weight of the edge connecting word $p$ and word $q$. Note that $\mathbold{A}$ does not contain self-loop, and we add a self-loop for each node.

\xhdr{GCN Layer} After top-$k$ selection on each attention score matrix $\mathbold{A}_{score}^i$ ($\mathbold{A}_{score}^i$ is not fully connected anymore), we apply a one-layer GCN and get updated node features as follows:
\begin{flalign}
\mathbold{\hat{H}}^{(l,i)}&=\sigma(\mathbold{A}^i\mathbold{\hat{H}}^{(l-1)}\mathbold{W}^i) + \mathbold{\hat{H}}^{(l-1)}\mathbold{W}^i\\
\mathbold{\hat{H}}^{(l)}&=\mathbin\Vert_{i=1}^{L}\mathbold{{\hat{H}}}^{(l,i)}
\end{flalign}
where $\mathbold{\hat{H}}^{(l)} \in \mathbb{R}^{n \times d_h}$ is the output of the $l$-th \textit{SA-GCN} block and composed by the concatenation of $\mathbold{\hat{H}}^{(l,i)} \in \mathbb{R}^{n \times d_{head}}$ of $i$-th head, $\mathbold{\hat{H}}^{(0)} \in \mathbb{R}^{n \times d_h}$ is the input of the first \textit{SA-GCN} block and comes from the GCN layer operating on the dependency tree, $\mathbold{{A}^i}$ is the top-$k$ score matrix of $i$-th head, $\mathbold{W}^i \in \mathbb{R}^{d_h \times d_{head}}$ denotes as the learnable weight matrix, and $\sigma$ refers to $ReLU$ activation function. The \textit{SA-GCN} block can be applied multi times if needed.

\subsection{Classifier}
We extract the aspect term node feature from $\mathbold{\hat{H}}_{o}$, which is the output of the last \textit{SA-GCN} block, and conduct the average pooling to obtain $\mathbold{\hat{h}}_t \in \mathbb{R}^{1 \times d_{h}}$. Then we feed it into a two-layer MLP to calculate the final classification scores $\hat{\mathbold{y}}_s$:
\begin{flalign}
\hat{\mathbold{y}}_s&=softmax({\mathbold{W}_{2}\sigma({\mathbold{W}_{1}\mathbold{\hat{h}}_{t}^T})})
\end{flalign}
where $\mathbold{W}_{2} \in \mathbb{R}^{C \times d_{out}}$ and $\mathbold{W}_{1} \in \mathbb{R}^{d_{out} \times d_{h}}$ denote the learnable weight matrix, $C$ is the sentiment class number,
%which is 3 in our case, 
and $\sigma$ refers to $ReLU$ activation function. 
We use cross entropy as the sentiment classification loss function:
\begin{flalign}
L_{s}&=-\sum_{c=1}^{C}\mathbold{y}_{s,c}\log\hat{\mathbold{y}}_{s,c}+\lambda {\left\lVert\theta\right\rVert}^2
\end{flalign}
where $\lambda$ is the coefficient for L2-regularization, $\theta$ denotes the parameters that need to be regularized, $\mathbold{y}_s$ is the true sentiment label.

\subsection{Opinion Extractor}
The opinion extraction shares the same input encoder, i.e. the \textit{SA-GCN} as sentiment classification. Therefore we feed the output of \textit{SA-GCN} to a linear-chain Conditional Random Field (CRF)~\cite{lafferty2001conditional}, which is the opinion extractor. Specifically, based on the \textit{SA-GCN} output $\mathbold{\hat{H}}_{o}$, the output sequence $\mathbold{y}_o = [y_1, y_2, \cdots, y_n]$ ($y_{i}\in \{B, I, O\}$) is predicted as:
\begin{flalign}
p(\mathbold{y}_o|\mathbold{\hat{H}}_{o})&=\frac{exp(s(\mathbold{\hat{H}}_{o},\mathbold{y}_o))}{\sum_{{\mathbold{y}\prime}_o \in Y}exp(s(\mathbold{\hat{H}}_{o},{\mathbold{y}\prime}_o))} \\
s(\mathbold{\hat{H}}_{o},\mathbold{y}_o)&=\sum_{i}^{n}(\mathbold{T}_{y_{i-1}, y_i}+\mathbold{P}_{i,y_i})\\
\mathbold{P}_{i} & = \mathbold{W}_{o}\mathbold{\hat{H}}_{o}[i]+\mathbold{b}_{o}
\end{flalign}
where $Y$ denotes the set of all possible tag sequences, $\mathbold{T}_{y_{i-1}, y_i}$ is the transition score matrix, $\mathbold{W}_{o}$ and $\mathbold{b}_{o}$ are learnable parameters. We apply Viterbi algorithm in the decoding phase. And the loss for opinion extraction task is defined as:
\begin{flalign}
L_{o}&=-log(p(\mathbold{y}_o|\mathbold{\hat{H}}_{o}))
\end{flalign}

Finally, the total training loss is:
\begin{flalign}
L&=L_{s}+\alpha{L_{o}}
\end{flalign}
where $\alpha > 0$ represents the weight of opinion extraction task.
%%%%%%%%%%%%%%%%%%%%%%%%%%%%%%%%%%%%%%%%%%%%%%%%%%%%%%%
\begin{table}[!ht]
\centering
\begin{adjustbox}{max width=0.48\textwidth}
\begin{tabular}{ c c c c c c c }
\hline
\multirow{2} {*}{Dataset} & \multicolumn{2}{c}{Positive} & \multicolumn{2}{c}{Neutral} & \multicolumn{2}{c}{Negative}\\
\cline{2-7}
& Train & Test & Train & Test & Train & Test\\
\hline
14Lap & 991 & 341 & 462 & 169 & 867 & 128\\
14Rest & 2164 & 728 & 633 & 196 & 805 & 196\\
15Rest & 963 & 353 & 36 & 37 & 280 & 207\\
16Rest & 1324 & 483 & 71 & 32 & 489 & 135\\
% Twitter & 1561 & 173 & 3127 & 346 & 1560 & 173 \\
\hline
\end{tabular}
\end{adjustbox}
\caption{Statistics of Datasets.}
\label{table:data}
\end{table}
\vspace{-12pt}
%%%%%%%%%%%%%%%%%%%%%%%%%%%%%%%%%%%%%%%%%%%%%%%%%%%%%%%
\section{Experiments}\label{sec:experiments}
\xhdr{Data Sets} We evaluate our \textit{SA-GCN} model on four datasets: Laptop reviews from SemEval 2014 Task 4 (14Lap), Restaurant reviews from SemEval 2014 Task 4, SemEval 2015 Task 12 and SemEval 2016 Task 5 (14Rest, 15Rest and 16Rest). %We randomly sample 5\% training data as development set and use the remaining 95\% for training.
We remove several examples with ``conflict'' labels.  The statistics of these datasets are listed in Table \ref{table:data}.

\xhdr{Baselines} Since BERT\cite{devlin2018bert} model shows significant improvements over many NLP tasks, we directly implement \textit{SA-GCN} based on BERT and compare with following  BERT-based baseline models:
\begin{enumerate}[itemsep=1pt,leftmargin=10pt]
\item
\textbf{BERT-SPC}~\cite{song2019attentional} feeds the sentence and term pair into the BERT model and the BERT outputs are used for prediction.
\item
\textbf{AEN-BERT}~\cite{song2019attentional} uses BERT as the encoder and employs several attention layers.
% \item
% \textbf{RACL-BERT}~\cite{chen-qian-2020-relation} learns the relations among aspect-term sentiment classification, opinion extraction and aspect-term extraction tasks, to help each individual task via the multi-task learning.
\item
\textbf{TD-GAT-BERT}~\cite{huang2019syntax} utilizes GAT on the dependency tree to propagate features from the syntactic context.
\item
\textbf{DGEDT-BERT}~\cite{tang-etal-2020-dependency} proposes a mutual biaffine module to jointly consider the flat representations learnt from Transformer and graph-based representations learnt from the corresponding dependency graph in an iterative manner.
\item
\textbf{R-GAT+BERT}~\cite{wang2020relational} reshapes and prunes the dependency parsing tree to an aspect-oriented tree rooted at the aspect term, and then employs relational GAT to encode the new tree for sentiment predictions.
\end{enumerate}
% reshapes and prunes the dependency parsing tree to an aspect-oriented tree rooted at the aspect term, and then employs Relational GAT to encode the new tree for sentiment predictions.
%with \textit{SA-GCN}.

%In Table~\ref{table:results} we present results of the average and standard deviation numbers from seven runs of random initialization.

In our experiments, we present results of the average and standard deviation numbers from seven runs of random initialization. We use BERT-base model to compare with other published numbers. We implement our own \textit{BERT-baseline} by directly applying a classifier on top of BERT-base encoder,
\textit{BERT+2-layer GCN} and \textit{BERT+4-layer GCN} are models with 2-layer and 4-layer GCN respectively on dependency trees with the BERT encoder. 
\textit{BERT+SA-GCN} is our proposed \textit{SA-GCN} model 
% (2-layer GCN + 1 SA-GCN block) 
with BERT encoder.
\textit{Joint SA-GCN} refers to joint training of sentiment classification and opinion extraction tasks. \textit{Joint SA-GCN (Best)} denotes as the best performances of our \textit{SA-GCN} model from the seven runs.

% \textbf{1-layer GCN on DT + 2 SA-GCN blocks with BERT-large} replaces the encoder with the whole word masking BERT large model.
% \begin{enumerate}
% \itemsep0em
    % \item

\begin{table*}[!ht]
\centering
\resizebox{2.0\columnwidth}{!}{
\begin{threeparttable}
%\begin{adjustbox}{max width=0.85\textwidth}
\begin{tabularx}{1.65\textwidth}{c|c c c c c c c c c}
\hline
\multirow{2} {*}{Category}  &\multirow{2} {*}{Model}  & \multicolumn{2}{c}{14Rest} & \multicolumn{2}{c}{14Lap} & \multicolumn{2}{c}{15Rest} & \multicolumn{2}{c}{16Rest} \\
\cline{3-10}
%& \multicolumn{2}{c}{Twitter}\\ 
 & & Acc & Macro-F1 & Acc & Macro-F1 & Acc & Macro-F1 & Acc & Macro-F1 \\
%& Acc & Macro-F1 \\
\hline
\multirow{2} {*}{BERT} &BERT-SPC
%\cite{song2019attentional} 
& 84.46 & 76.98 & 78.99 & 75.03 & - & - & - & - \\
%& 73.6 & 72.1 \\
&AEN-BERT
%\cite{song2019attentional} 
& 83.12 & 73.76 & 79.93 & 76.31 & - & - & - & -  \\
% &RACL-BERT
% %\cite{chen-qian-2020-relation} 
% & - & 81.61 & - & 73.91 & - & 74.91 & - & - \\ 
%& \textbf{74.7} & \textbf{73.1} \\
\hline
% \multirow{1} {*}{BERT+DT\tnote{$\star$}} &SDGCN-BERT\cite{zhaoa2019modeling} & 83.6 & 76.5 & 81.4 & 78.3 & - & -  \\
%& - & -\\
\multirow{2} {*}{BERT+DT\tnote{$\star$}} &TD-GAT-BERT
%\cite{huang2019syntax} 
& 83.0 & - & 80.1 & - & - & - & - & -\\
&DGEDT-BERT
%\cite{tang-etal-2020-dependency} 
& 86.3 & 80.0 & 79.8 & 75.6 & 84.0 & \textbf{71.0} & \textbf{91.9} & 79.0\\
% ASGCN\cite{zhang2019aspect} & 80.9 & 72.2 & 75.6 & 71.0 \\
% CDT\cite{sun2019aspect} & 82.3 & 74.0 & 77.2 & 73.0 \\
\hline
{BERT+RDT}{$^\diamond$} & R-GAT+BERT
%\cite{wang2020relational} 
& \textbf{86.60}  & \textbf{81.35}  & 78.21 & 74.07 & - & - & - & -  \\
\hline
\hline
%& 75.29 & 73.74
\multirow{6} {*}{Ours} & BERT-baseline & 85.56 $\pm$ 0.30 & 79.21 $\pm$ 0.45 & 79.57 $\pm$ 0.15 & 76.18 $\pm$ 0.31& 83.45 $\pm$ 1.13 & 69.29 $\pm$ 1.78 & 91.06 $\pm$ 0.44 & 78.58 $\pm$ 1.62\\
\cline{2-10}
&{BERT+2-layer GCN} & 85.78 $\pm$ 0.59 & 80.55 $\pm$ 0.90 & 79.72 $\pm$ 0.31 & 76.31 $\pm$ 0.35 & 83.71 $\pm$ 0.42 & 69.26 $\pm$ 1.63 & 91.23 $\pm$ 0.25 & 79.29 $\pm$ 0.51 \\
& BERT+4-layer GCN & 85.03 $\pm$ 0.64 & 78.90 $\pm$ 0.75 & 79.57 $\pm$ 0.15 & 76.23 $\pm$ 0.49 & 83.48 $\pm$ 0.33 & 68.72 $\pm$ 1.08  & 91.02 $\pm$ 0.26 & 78.68 $\pm$ 0.50\\
% &ensemble of GCN and BERT  & 84.2 & 77.0 & 79.2 & 75.2 & - & -\\
%& 74.4 & 72.4 \\
&{BERT+\textit{SA-GCN}}{$^\dagger$} & 86.16 $\pm$ 0.23 & 80.54 $\pm$ 0.38 & \textbf{80.31} $\pm$ 0.47 & \textbf{76.99} $\pm$ 0.59 & \textbf{84.18} $\pm$ 0.29 & 69.42 $\pm$ 0.81 & 91.41 $\pm$ 0.39 & \textbf{80.39} $\pm$ 0.93\\
\cline{2-10}
&Joint \textit{SA-GCN}& 86.57 $\pm$ 0.81 & 81.14 $\pm$ 0.69 & \textbf{80.61} $\pm$ 0.32 & \textbf{77.12} $\pm$ 0.51 & \textbf{84.63} $\pm$ 0.33 & 69.1 $\pm$ 0.78 & 91.54 $\pm$ 0.26 & \textbf{80.68} $\pm$ 0.92\\
&Joint \textit{SA-GCN} (Best)& \textbf{87.68} & \textbf{82.45} & \textbf{81.03} & \textbf{77.71} & \textbf{85.26} & 69.71 & \textbf{92.0} & \textbf{81.86}\\
\hline
\end{tabularx}
% \end{adjustbox}
		\begin{tablenotes}
			 \item $\star$ DT: Dependency Tree; $\diamond$ RDT: Reshaped Dependency Tree.
  			 \item
  		% 	 $\ddagger$ BERT+\textit{SA-GCN}: 2-layer GCN+1 \textit{SA-GCN} block; 
  			 $\dagger$: Head-independent based top-$k$ Selection.
		\end{tablenotes}
\end{threeparttable}
}
\caption{Comparison of \textit{SA-GCN} with various baselines.}
\label{table:results}
\end{table*}

\begin{table*}[!ht]
% \centering
\resizebox{0.99\linewidth}{!}{%
\begin{tabular}{l|lll}
\hline
Sentence & Label & GCN & \textit{SA-GCN} \\
\hline
%I'm looking forward to going back soon and eventually trying most everything on the \textcolor{red}{menu}! & positive & neutral & positive \\
% The food is really good, \colorbox{steelblue}{however} I had the \textcolor{red}{soup} and it was \colorbox{lightblue}{not} \colorbox{skyblue}{tasty}. & negative & positive & negative\\
%\hline
% \colorbox{steelblue}{Had} \textcolor{red}{dinner} here on a Friday and \colorbox{skyblue}{the} food \colorbox{lightblue}{was} great. & neutral & positive & neutral\\
\shortstack[l]{Satay is one of those \colorbox{steelblue}{favorite} haunts on Washington where the service and \colorbox{skyblue}{\textcolor{red}{food}} \colorbox{lightblue}{is} always on the money.} & positive & neutral & positive \\
% \shortstack[l]{The food is just OKAY, and it's almost not worth going unless you're \\\colorbox{lightblue}{getting} the \textcolor{red}{pialla}, which is the only \colorbox{steelblue}{dish} that's really \colorbox{skyblue}{good}.} & positive & neutral & positive\\
\hline
And the fact that it comes with an \textcolor{red}{i5 processor} definitely \colorbox{steelblue}{speeds} \colorbox{lightblue}{things} \colorbox{skyblue}{up} & positive & neutral & positive\\
% \shortstack[l]{but, the filet mignon was not very good at all \textcolor{red}{\colorbox{lightblue}{cocktail} hour} \colorbox{skyblue}{includes} free \colorbox{steelblue}{appetizers} (nice non-sushi selection).} & positive & neutral & positive\\
% \shortstack[l]{Only suggestion is that you skip the \textcolor{red}{dessert}, \colorbox{skyblue}{it} was \colorbox{lightblue}{overpriced} and fell\\ \colorbox{steelblue}{short} on taste.} & negative & neutral & negative\\
% \textcolor{red}{food} & Although the restaurant itself is nice, \colorbox{lightblue}{I} prefer \colorbox{skyblue}{not} to go for the \colorbox{steelblue}{food}.\\
\hline
 I know real \textcolor{red}{Indian food} \colorbox{steelblue}{and} this \colorbox{skyblue}{was} \colorbox{lightblue}{n't} it. & negative & neutral & negative\\
 \hline
% Didn't seem like any effort was made to the \textcolor{red}{display and quality of the food}.
% & negative & positive & negative\\
% \hline
\end{tabular}%
}
\caption{Top-$k$ visualization: the darker the shade, the larger attention weight.}
\label{table:attention}
\end{table*}

\xhdr{Parameter Setting} During training, we set the learning rate to $10^{-5}$. The batch size is $4$.
We obtain dependency trees using the Stanford Stanza~\cite{qi2020stanza}.
% Stanford CoreNLP~\cite{manning2014stanford}. 
The dimension of BERT output $d_{B}$ is 768. The hidden dimensions are selected from $\{128, 256, 512\}$. 
We apply dropout~\cite{srivastava2014dropout} and the dropout rate range is $[0.1,0.4]$. The L2 regularization is set to $10^{-6}$. We use $1$ or $2$ \textit{SA-GCN} blocks in our experiments. We choose $k$ in top-$k$ selection module from $\{2,3\}$ to achieve the best performance.
For joint training, the weight range of opinion extraction loss is $[0.05, 0.15]$ \footnote{Our code will be released at the time of publication.}.
% we follow the same testing procedure as previous work did~\cite{wang2020relational}.
% We report the average results of seven runs with random initialization.
% \footnote{Our code will be released at the time of publication.}
% The coefficient rate $\lambda$ of L2 is $10^{-6}$. \footnote{We will release the code after paper review.}

% \xhdr{Evaluation Settings} The evaluation metrics are accuracy and Macro-F1.

\subsection{Experimental Results}
We present results of the \textit{SA-GCN} model in two aspects: classification performance and qualitative case study.

\xhdr{Classification} Table \ref{table:results} shows comparisons of \textit{SA-GCN} with other baselines in terms of classification accuracy and Macro-F1. From this table, we observe that:
% \jing{worth mention our BERT baseline got really high performance compared to other papers, especially on laptop data.}
\textit{SA-GCN} achieves the best average results on 14Lap, 15Rest and 16Rest datasets, and obtains competitive results on 14Rest dataset. The joint training of sentiment classification and opinion extraction tasks further boosts the performances on all datasets.
% outperforms the current state-of-the-art by 1.3\% and 0.3\% in terms of accuracy, and 2.7\% and 0.5\% in terms of Macro-F1. 
% These results are the new state-of-the-art results (to our best knowledge). 
%On the Twitter dataset, \textit{SA-GCN} shows competitive results as well. 

Specifically, \textit{BERT+2-layer GCN} outperforms \textit{BERT-baseline}, which proves the benefit of using syntax information.
\textit{BERT+4-layer GCN} is actually worse than \textit{BERT+2-layer GCN}, which shows that more GCN layers do not bring additional gain.

Our \textit{BERT+SA-GCN} model 
% with 2-layer GCN + 1 SA-GCN block 
further outperforms the \textit{BERT+2-layer GCN} model. Because the \textit{SA-GCN} block allows aspect terms to directly absorb the information from the most important context words that are not reachable within two hops in the dependency tree. 

Besides, introducing the opinion extraction task provides more supervision signals for the top-$k$ selection module, which benefits the sentiment classification task.

\xhdr{Qualitative case Study} To show the efficacy of the \textit{SA-GCN} model on dealing long-hops between aspect term and its opinion words, we demonstrate three examples as shown in Table \ref{table:attention}. 
These sentences are selected from test sets of 14Lap and 14Rest datasets and predicted correctly by the \textit{SA-GCN} model but wrongly by \textit{BERT+2-layer GCN}. 
The important thing to note here, our \textit{SA-GCN} model could provide explanation about the prediction according to the learned attention weights, while the GCN based model (\textit{BERT+2-layer GCN} denoted as ``GCN'' in Table \ref{table:attention}) cannot.
Aspect terms are colored red. Top-3 words with the largest attention weights towards the aspect term are shaded. The darker the shade, the large attention weight. 

In all three examples the aspect terms are more than three hops away from essential opinion words\footnote{See their dependency trees in supplemental material.}, thus \textit{BERT+2-layer GCN} model cannot learn the interactions between them within two layers, while \textit{SA-GCN} model overcomes the distance limitation and locates right opinion words.
% In the first example, the distance between aspect term ``food'' and opinion words ``favorite'' and ``on the money'' is more than three hops. \textit{SA-GCN} overcomes the distance limit and locates the right opinion word.
% In the second example, there are three hops between the apsect term ``i5 processor'' and opinion words ``speeds things up''
% the label of the aspect term ``pialla'' is positive, however it is predicted as neutral by the GCN model. The \textit{SA-GCN} model is able to focus on the crucial opinion word ``good'' and makes the right prediction.
% In the third example, the aspect term ``dessert'' is more than three-hops away from opinion words ``overpriced'' and  ``short on taste'' on the dependency tree. The proposed \textit{SA-GCN} overcomes the hop distance limitation and directly finds the right opinion words. 
%The fourth example contains two terms ``restaurant'' and ``food'' with conflict sentiment polarity, the SA-GCN model is not distracted by the misleading word ``nice'' and locates the related context words for ``food''.

\begin{table}[!h]
\vspace{-10pt}
\centering
\begin{adjustbox}{max width=0.5\textwidth}
\begin{tabular}{l |c  c  c  c}
\hline
\multirow{2} {*}{Model} & 14Rest & 14Lap & 15Rest & 16Rest\\
\cline{2-5}
& F1 & F1 & F1 & F1 \\
\hline
IOG & 80.24 & 71.39 & 73.51  & 81.84 \\
ASTE  & 83.15 & 76.03 & 78.02 & 83.73 \\
\hline
Joint \textit{SA-GCN} & \textbf{83.72} $\pm$ 0.51 & \textbf{76.79} $\pm$ 0.33 & \textbf{80.99} $\pm$ 0.43 & \textbf{83.83} $\pm$ 0.50 \\
\hline
\end{tabular}
\end{adjustbox}
\caption{Opinion extraction results.}
\label{tab:opinion}
\end{table}

\begin{table*}[!h]
\centering
\resizebox{2.1\columnwidth}{!}{
\begin{threeparttable}
\begin{tabularx}{1.45\textwidth}{l c c c c c c c c}
\hline
% \multirow{2} {*}{} &
\multirow{2} {*}{Model}  & \multicolumn{2}{c}{14Rest} & \multicolumn{2}{c}{14Lap} & \multicolumn{2}{c}{15Rest} & \multicolumn{2}{c}{16Rest}\\
\cline{2-9}
%& \multicolumn{2}{c}{Twitter}\\ 
& Acc & Macro-F1 & Acc & Macro-F1 & Acc & Macro-F1 & Acc & Macro-F1 \\
%& Acc & Macro-F1 \\
\hline
% 2-layer GCN  & 85.18 & 79.08 & 79.47 & 75.98 & - & - & - & - \\
% 1-layer GCN+1 \textit{SA-GCN} block & 86.25 & 80.54 & 79.94 & 76.65 & - & - & - & - \\
% \hline
\textit{SA-GCN} (head-ind) & \textbf{86.16} $\pm$ 0.23 & \textbf{80.54} $\pm$ 0.38 & \textbf{80.31} $\pm$ 0.47 & \textbf{76.99} $\pm$ 0.59 & \textbf{84.18} $\pm$ 0.29 & 69.42 $\pm$ 0.81 &\textbf{91.41}$\pm$ 0.39 & \textbf{80.39} $\pm$ 0.93 \\
\hline
\textit{SA-GCN} w/o top-k & 85.06 $\pm$ 0.68 & 78.88 $\pm$ 0.83 & 79.96 $\pm$ 0.14 & 76.64 $\pm$ 0.58 & 83.15 $\pm$ 0.41 & 68.74 $\pm$ 1.48 & 90.92 $\pm$ 0.45 & 78.18 $\pm$ 0.71 \\
\textit{SA-GCN} (head-dep)\tnote{$\diamond$} & 85.41 $\pm$ 0.21 & 79.19 $\pm$ 0.68 & 80.17 $\pm$ 0.55& 76.83 $\pm$ 0.59 & 83.68 $\pm$ 0.54 & 68.81 $\pm$ 1.39 & 91.01 $\pm$ 0.40 & 78.88 $\pm$ 1.04\\
\hline
\end{tabularx}
% \end{adjustbox}
		\begin{tablenotes}
			 \item [$\diamond$] head-dep: head-dependent based top-$k$ selection.
		\end{tablenotes}
\end{threeparttable}
}
\caption{Ablation study of \textit{SA-GCN}.}
\label{table:ablation}
\end{table*}
\vspace{-12pt}

\xhdr{Opinion Extraction} Table \ref{tab:opinion} shows the results of the opinion extraction task under the joint training setting. The reported numbers are obtained by averaging F1 of seven runs. In each run, the selected opinion F1 is generated from the best sentiment classification checkpoint. We compare our model with two baselines: \textbf{IOG}~\cite{fan2019target} encodes the aspect term information into context by an Inward-Outward
LSTM to find the corresponding opinion words.
\textbf{ASTE}~\cite{peng2020knowing} utilizes a GCN module to learn the mutual dependency relations between different words and to guide opinion term extraction.
As shown in this table, the joint \textit{SA-GCN} model outperforms two baseline models on all datasets, which demonstrates that the sentiment classification task is helpful for opinion extraction task as well.

\subsection{Model Analysis}
We further analyze our \textit{SA-GCN} model from two perspectives: ablation study and sentence length analysis.

\xhdr{Ablation Study} To demonstrate effectiveness of different modules in \textit{SA-GCN}, we conduct ablation studies in Table \ref{table:ablation}.
% by the following settings:
% \begin{itemize}[leftmargin=10pt]
% \itemsep0em
%     \item \textbf{2-layer GCN} applies the 2-layer GCN on dependency trees based on BERT-based encoder.
%     \item \textbf{1-layer GCN+1 SA-GCN block} utilizes 1-layer GCN on dependency trees and then employs one SA-GCN block.
%     % \item \textbf{2 SA-GCN blocks} directly applies two SA-GCN blocks, and removes the GCN over dependency trees.
%     \item \textbf{SA-GCN w/o top-k} ablates the top-k selection module in the SA-GCN block.
%     \item \textbf{SA-GCN(head-dep)} and \textbf{SA-GCN(head-ind)} refer to head-dependent and head-independent selection respectively. 
% \end{itemize}
%shows the results of ablation study. 
% Notice that both \textit{SA-GCN w/o top-k} and \textit{SA-GCN(head-dep)} are implemented under the 2-layer GCN + 1 \textit{SA-GCN} block setting. 
From this table, we observe that: 
\begin{enumerate}[itemsep=0pt,leftmargin=10pt]
    % \item \xhdr{Effect of \textit{SA-GCN}} Compared with \textit{2-layer GCN}, our \textit{1-layer GCN+1 SA-GCN block} model improves on all three datasets, which shows that \textit{SA-GCN} block is beneficial to this task.
    
    %This is because the proposed SA-GCN block allows the aspect terms to directly absorb the information from the most important context words that are not reachable within two hops in the dependency tree. Even with parsing errors from the dependency tree, the joint training of GCN layer over the dependency tree followed by the SA-GCN blocks is able to correct the wrong context from parsing errors. Therefore the SA-GCN model is much more effective than \textit{ensemble of 2-layer GCN on DT and BERT}.
    %a simple ensemble of two models: GCN model over the dependency tree and a self-attention sequence model.

% We give an example in Figure \ref{fig:dt} as a case study: the example is labeled as ``positive'' towards the aspect term ``content creation'', however it's classified as ``neutral'' by \textit{2 SA-GCN blocks} model without dependency trees. In contrast, with the help of syntactic information introduced by the dependency tree, ``content creation'' is directly linked with the sentiment context word ``reliable'', \textit{1-layer GCN on dependency trees + 1 SA-GCN block} makes a correct prediction.
    \item \xhdr{Effect of Top-k Selection} To examine the impact the top-k selection, we present the result of \textit{SA-GCN w/o top-k} in Table \ref{table:ablation}. We can see that without top-k selection, both accuracy and macro-F1 decrease on all datasets. This observation proves that the top-k selection helps to reduce the noisy context and locate top important opinion words. We also conduct the effect of the hyper-parameter $k$ and the block number $N$ on \textit{SA-GCN} under head-independent and head-dependent selection respectively (see the supplemental material).
    
    \item \xhdr{Effect of Head-independent and Head-dependent Selection} As shown in the last row in Table \ref{table:ablation}, head-independent selection achieves better results than head-dependent selection. This is because the mechanism of head-independent selection is similar to voting. By summing up the weight scores from each head, context words with higher scores in most heads get emphasized, and words that only show importance in few heads are filtered out. Thus all heads reach to an agreement and the top-$k$ context words are decided. However for head-dependent selection, each head selects different top-$k$ context words, which is more likely to choose certain unimportant context words and introduce noise to the model prediction.
\end{enumerate}

%%%%%%%%%%%%%%%%%%%%%%%%%%%%%%%%%%%%%%%%%%%%%%
\begin{figure}[!h]
\centering
\begin{subfigure}[h]{.5\textwidth}
\centering
\includegraphics[width=0.7\linewidth, height=4.5cm]{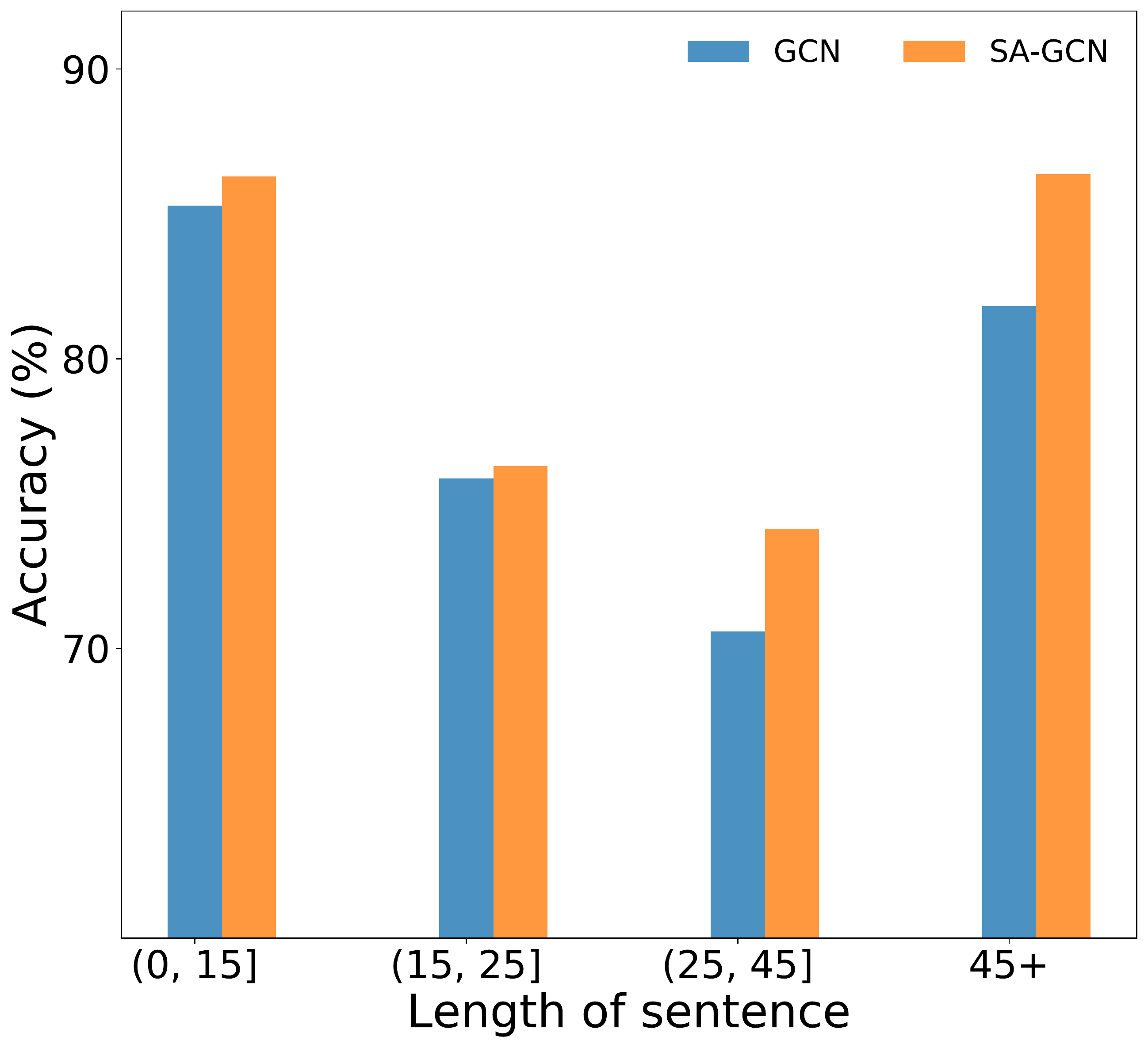}
\caption{Length analysis on 14Lap.}
% \label{fig:lengthLap}
\end{subfigure}
\begin{subfigure}[h]{.5\textwidth}
\centering
\includegraphics[width=0.7\linewidth, height=4.5cm]{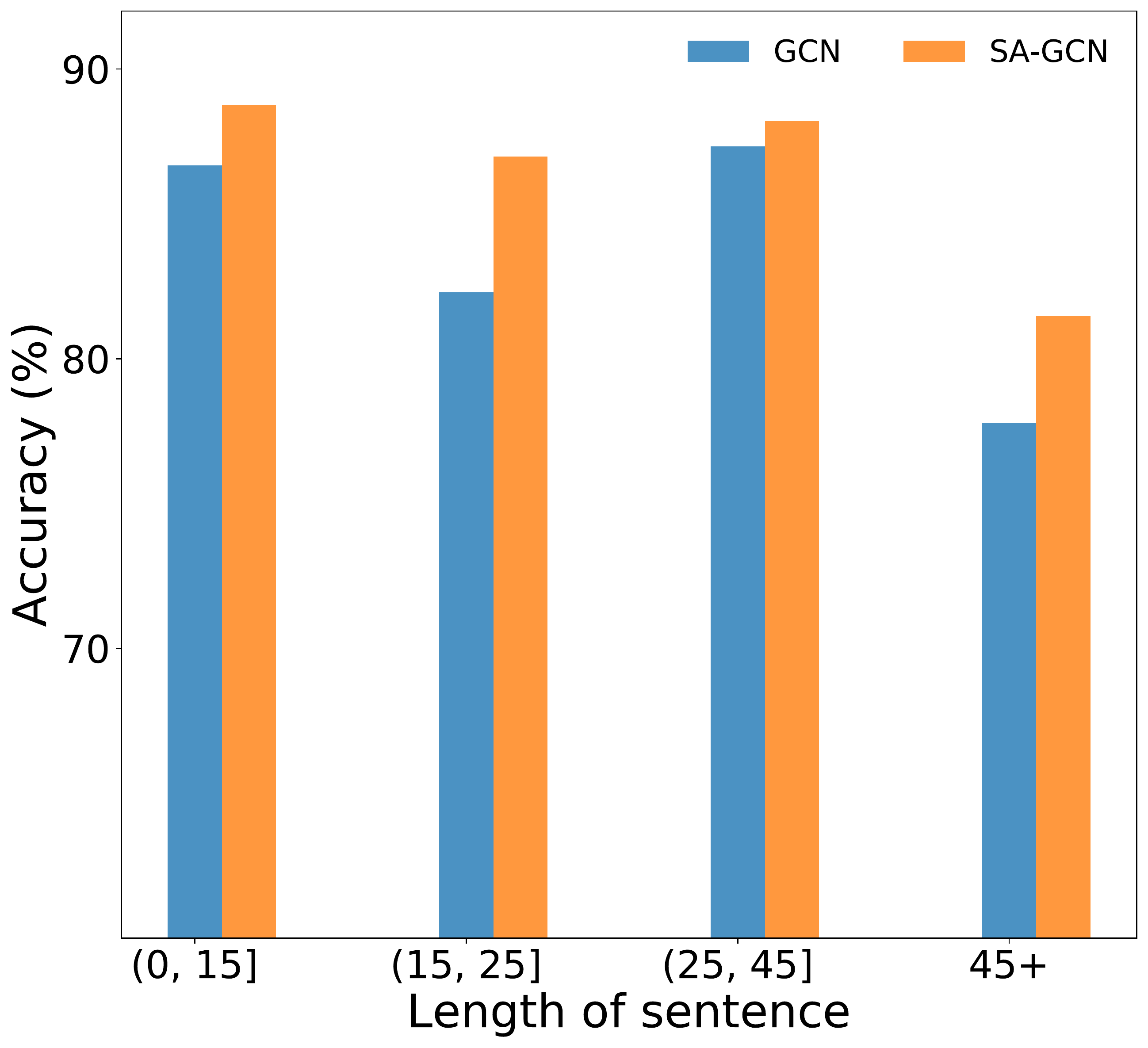}  
\caption{Length analysis on 14Rest.}\label{fig:block}
% \label{fig:lengthRest}
\end{subfigure}
\caption{Sentence length analysis on 14Lap and 14Rest.}
\label{fig:lengthanalysis}
\end{figure}
%%%%%%%%%%%%%%%%%%%%%%%%%%%%%%%%%%%%%%%%%%%%%%

\xhdr{Sentence Length Analysis} To quantify the ability of our \textit{SA-GCN} model dealing with long-distance problem, we conduct sentence length analysis on 14Lap and 14Rest datasets. The assumption is that the longer the sentence, the more likely the long-distance problem occurs. The results are showed in Figure \ref{fig:lengthanalysis}. We measure the sentiment classification accuracy of \textit{BERT+2-layer GCN} (denotes as GCN in Figure \ref{fig:lengthanalysis}) and \textit{BERT+SA-GCN} models under different sentence lengths. We observe that \textit{SA-GCN} achieves better accuracy than GCN across all length ranges and is more advantageous when sentences are longer. To some extent, the results prove effectiveness of \textit{SA-GCN} in dealing with long-distance problem.  

\section{Conclusions}
We propose a selective attention based GCN model for the aspect-level sentiment classification task. We first encode the aspect term and context words by pre-trained BERT to capture the interaction between them, then build a GCN on the dependency tree to incorporate syntax information. In order to handle the long distance between aspect terms and opinion words, we use the selective attention based GCN block, to select the top-$k$ important context words and employ the GCN to integrate their information for the aspect term representation learning. Further, we adopt opinion extraction problem as an auxiliary task to jointly train with sentiment classification task. We conduct experiments on several SemEval datasets. The results show that \textit{SA-GCN} outperforms previous strong baselines and achieves the new state-of-the-art results on these datasets.
\bibliographystyle{acl_natbib}
\bibliography{anthology,eacl2021}

\clearpage
\onecolumn
\section{Supplemental Material}
\begin{figure*}[!h]
\centering
\begin{subfigure}{0.9\textwidth}
\centering
\includegraphics[width=\linewidth]{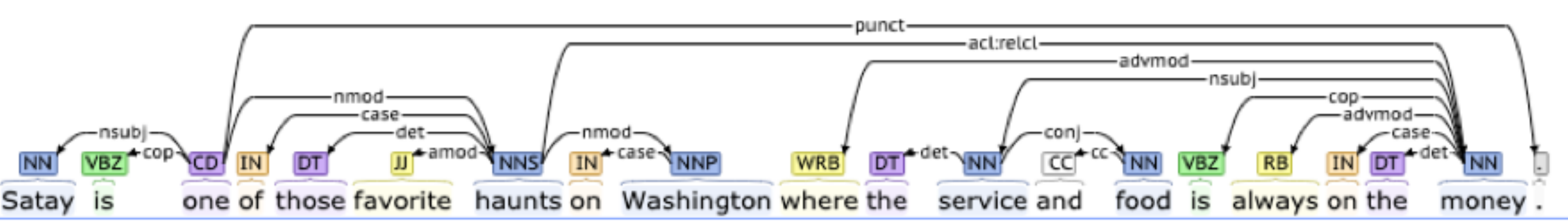}
\caption{case 1}
\label{fig:case1}
\end{subfigure}
\begin{subfigure}{1.0\textwidth}
\includegraphics[width=\linewidth]{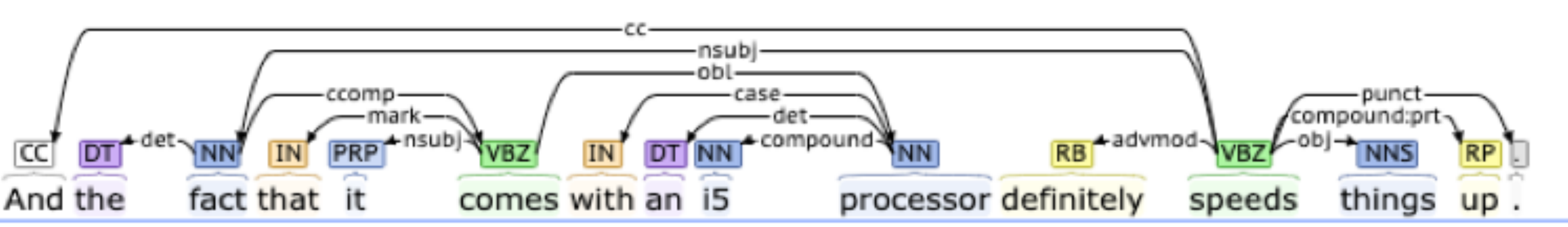}  
\caption{case 2}\label{fig:case2}
\end{subfigure}
\begin{subfigure}{1.0\textwidth}
\includegraphics[width=\linewidth]{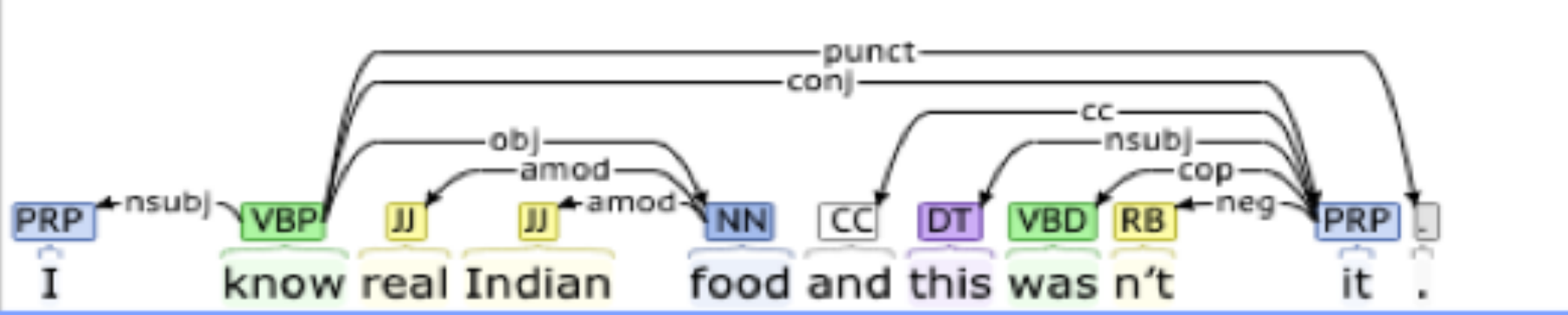}  
\caption{case 3}\label{fig:case3}
\end{subfigure}
\caption{Dependency trees of case study.}
\label{fig:casestudy}
\end{figure*}

\xhdr{Dependency Trees} The dependency trees of 3 cases in Table \ref{table:attention} are presented in Figure \ref{fig:casestudy}. The dependency trees are obtained from Stanford Stanza toolkit. As shown in the figure, in the first example, the aspect term ``food'' is four hops away from the opinion words ``favorite'' and ``on the money''. In the second and third example, there are also three-hops distance between aspect terms and opinion words. All three cases are correctly predicted by our \textit{SA-GCN} model, but wrongly by the 2-layer GCN model.

\begin{figure}[!ht]
\centering
\begin{subfigure}{.495\textwidth}
\centering
\includegraphics[width=\linewidth,]{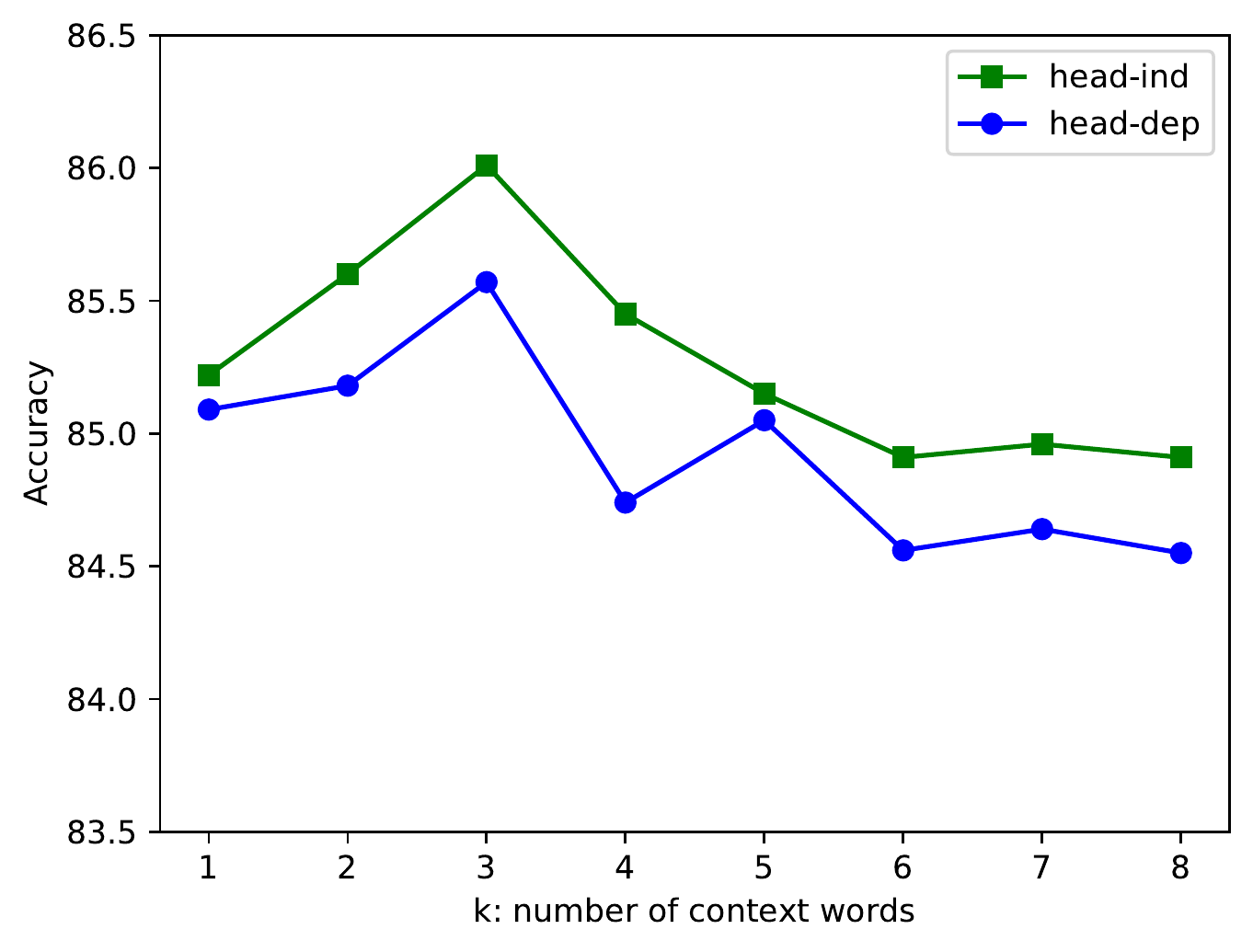}
\caption{Impact of $k$}
\label{fig:top-k}
\end{subfigure}
\begin{subfigure}{.495\textwidth}
\includegraphics[width=\linewidth]{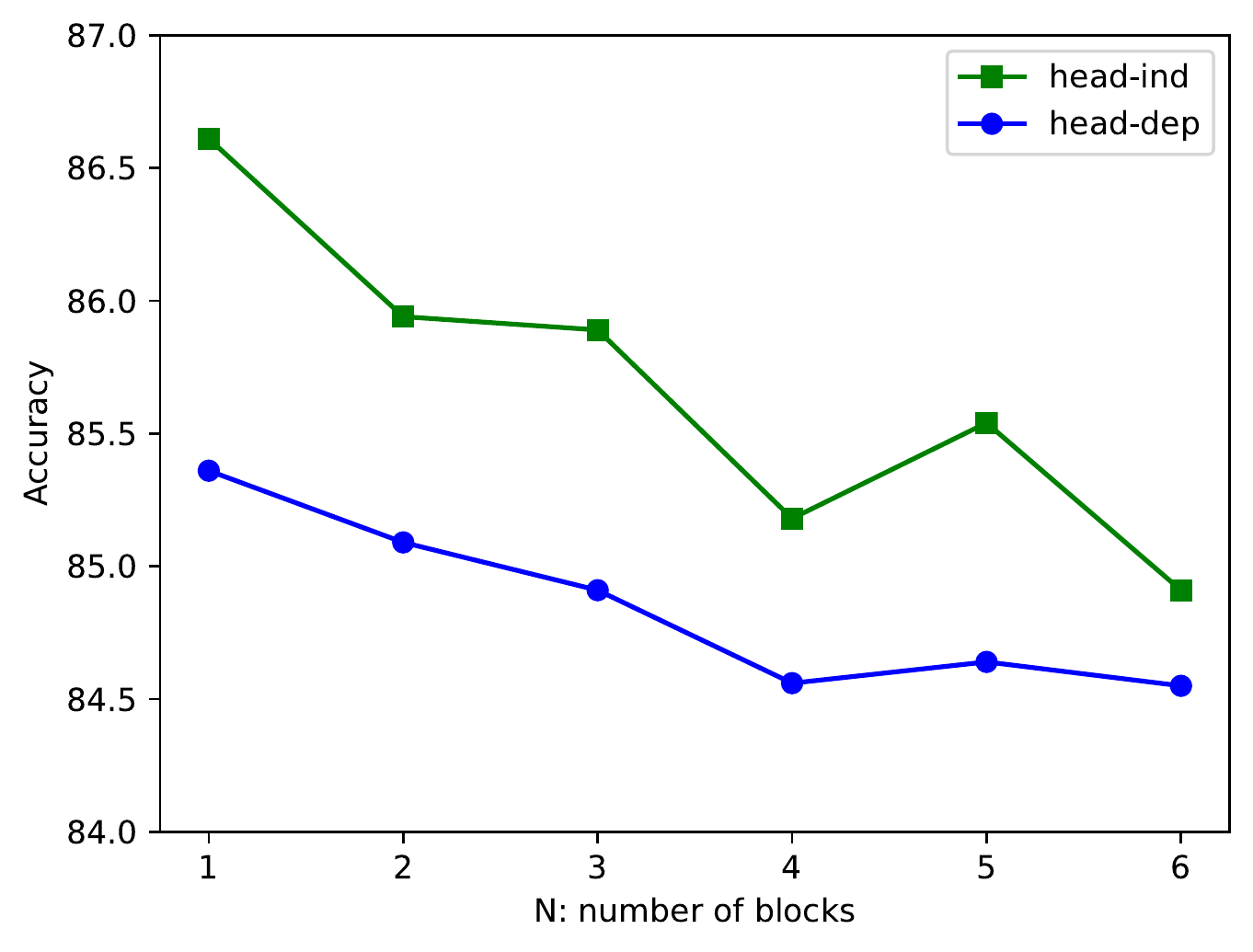}  
\caption{Impact of block numbers}\label{fig:block}
\end{subfigure}
\caption{Impact of $k$ and block numbers on \textit{SA-GCN} over Restaurant dataset.}
\label{fig:impactanalysis}
\vspace{-10pt}
\end{figure}

\xhdr{Hyper-parameter Analysis} We examine the effect of the hype-parameter $k$ and the block number $N$ on our proposed model under head-independent and head-dependent selection respectively. Figure \ref{fig:impactanalysis} shows the results on 14Rest dataset.
%%%%%%%%%%%%%%%%%%%%%%%%%%%%%%%%%%%%%%%%%%%%%%
%%%%%%%%%%%%%%%%%%%%%%%%%%%%%%%%%%%%%%%%%%%%%%
\begin{enumerate}[leftmargin=10pt]
\itemsep0em
\item
\xhdr{Effect of Hyper-parameter $k$} From Figure \ref{fig:top-k}, we observe that: 1) the highest accuracy appears when $k$ is equal to 3. As $k$ becomes bigger, the accuracy goes down. The reason is that integrating information from too many context words could introduce distractions and confuse the representation of the current word. 
2) Head-independent selection performs better than head-dependent selection as $k$ increases. As mentioned before, compared with head-independent, head-dependent selection might have more than $k$ context words contribute to the aggregation and introduce some noise.
\item\xhdr{Effect of Block Number} Figure \ref{fig:block} shows the effect of different number of \textit{SA-GCN} blocks. As the block number increases, the accuracy decreases for both head-independent and head-dependent selection. A single \textit{SA-GCN} block is sufficient for selecting top-$k$ important context words.
Stacking multiple blocks introduces more parameters and thus would lead to over-fitting with such a small amount of training data. This might be the reason why stacking multiple blocks is not helpful.
For our future work we plan to look into suitable deeper GNN models that are good for this task.
\end{enumerate}

\end{document}